\documentclass[pmlr]{jmlr}

\RequirePackage{graphicx}
 \usepackage{booktabs}
\usepackage{longtable}
\usepackage{capt-of}
\usepackage{microtype}
\usepackage{enumitem}
\usepackage{multirow}
\usepackage{array}
\usepackage{xcolor}
\usepackage{pgfplots}
\pgfplotsset{compat=1.18}
\usepackage{hyperref}
\hypersetup{
  pdftitle={SHIFT-M3: Pre-fusion Alignment-based Consistency Screening for Multimodal ECG Record Integrity},
  pdfauthor={Md Ashik Khan and Md Nahid Siddique}
}
\usepackage{float}
\makeatletter
\def\set@curr@file#1{\def\@curr@file{#1}} 
\makeatother
\usepackage[load-configurations=version-1]{siunitx} 

\theorembodyfont{\upshape}
\theoremheaderfont{\scshape}
\theorempostheader{:}
\theoremsep{\newline}

\jmlrproceedings{PMLR}{Preprint}
\jmlryear{2026}
\jmlrworkshop{Accepted at MLHC 2026}

\title[SHIFT-M3 for ECG Record Integrity]{SHIFT-M3: Pre-fusion Alignment-based Consistency Screening for Multimodal ECG Record Integrity}

\author{\Name{Md Ashik Khan}
       \Email{aasshhik98@gmail.com}\\
       \addr Department of Computer Science and Engineering\\
       Indian Institute of Technology Kharagpur\\
       Kharagpur, India
       \AND
       \Name{Md Nahid Siddique}
       \Email{msidd040@fiu.edu}\\
       \addr Knight Foundation School of Computing and Information Sciences\\
       Florida International University\\
       Miami, FL, USA}

\newcommand{\typeione}{97.6}
\newcommand{\typeithree}{90.3}
\newcommand{\hardneg}{97.7}
\newcommand{\typeiauroc}{0.996}
\newcommand{\typeiiiauroc}{0.974}
\newcommand{\hardnegauroc}{0.996}
\newcommand{\typeiifpr}{87.0}

\newcommand{\wordtypeione}{84.4}
\newcommand{\wordtypeithree}{24.4}
\newcommand{\wordtypeiifpr}{50.8}
\newcommand{\wordtypeiauroc}{0.965}
\newcommand{\wordtypeiiiauroc}{0.798}
\newcommand{\wordhardneg}{84.1}
\newcommand{\wordhardnegauroc}{0.965}

\newcommand{\chartypeione}{45.7}
\newcommand{\chartypeithree}{13.6}
\newcommand{\chartypeiifpr}{26.0}
\newcommand{\chartypeiauroc}{0.904}
\newcommand{\chartypeiiiauroc}{0.729}
\newcommand{\charhardneg}{46.0}
\newcommand{\charhardnegauroc}{0.903}

\newcommand{\logregtypeione}{89.1}
\newcommand{\logregtypeithree}{32.5}
\newcommand{\logregtypeiifpr}{58.1}
\newcommand{\logregtypeiauroc}{0.973}
\newcommand{\logregtypeiiiauroc}{0.824}
\newcommand{\logreghardneg}{89.2}
\newcommand{\logreghardnegauroc}{0.973}

\newcommand{\sweeptypeione}{97.21}
\newcommand{\sweeptypeithree}{89.13}
\newcommand{\sweephardneg}{97.32}
\newcommand{\sweeptypeiaurocfull}{0.995}
\newcommand{\sweeptypeiiiaurocfull}{0.970}
\newcommand{\sweeptypeiifpr}{84.13}
\newcommand{\sweeplambda}{1.0}
\newcommand{\sweepmargin}{0.5}
\newcommand{\agglambda}{2.0}
\newcommand{\aggmargin}{0.5}
\newcommand{\aggtypeione}{92.79}
\newcommand{\aggtypeithree}{77.44}
\newcommand{\agghardneg}{93.06}
\newcommand{\aggtypeiifpr}{73.07}
\newcommand{\aggtypeiauroc}{0.984}
\newcommand{\aggtypeiiiauroc}{0.936}

\newcommand{\maskedtypeione}{83.3}
\newcommand{\maskedtypeiauroc}{0.960}

\newcommand{\minilmtypeiauroc}{0.729}
\newcommand{\minilmtypeiiiauroc}{0.527}
\newcommand{\minilmhardnegauroc}{0.731}
\newcommand{\minilmtypeiifpr}{9.9}
\newcommand{\biocberttypeiauroc}{0.555}
\newcommand{\biocberttypeiiiauroc}{0.500}
\newcommand{\biocberthardnegauroc}{0.556}
\newcommand{\biocberttypeiifpr}{7.3}

\newcommand{\textspliceauroc}{0.927}
\newcommand{\textsplicemaskedauroc}{0.796}
\newcommand{\textsplicetprfive}{73.7}
\newcommand{\textsplicetprfivemasked}{31.5}
\newcommand{\maskedhardnegauroc}{0.967}

\newcommand{\frontierOneTI}{96.6}
\newcommand{\frontierOneTIII}{63.0}
\newcommand{\frontierOneHN}{96.8}
\newcommand{\frontierOneTII}{77.2}
\newcommand{\frontierFiveTI}{97.7}
\newcommand{\frontierFiveTIII}{73.7}
\newcommand{\frontierFiveHN}{97.8}
\newcommand{\frontierFiveTII}{82.8}
\newcommand{\frontierTenTI}{98.5}
\newcommand{\frontierTenTIII}{76.6}
\newcommand{\frontierTenHN}{98.4}
\newcommand{\frontierTenTII}{85.4}
\newcommand{\frontierTwentyTI}{99.7}
\newcommand{\frontierTwentyTIII}{84.3}
\newcommand{\frontierTwentyHN}{99.5}
\newcommand{\frontierTwentyTII}{90.3}

\begin{document}

\maketitle

\begin{abstract}
Multimodal clinical AI typically assumes that the waveform, report, metadata,
and downstream predictions attached to a record belong to the same patient.
In practice, linkage failures can silently assemble individually plausible but
cross-patient components, creating a safety problem that standard predictive
models are not designed to detect.
We study this problem as \emph{multimodal record integrity triage}: given an
assembled record, should its modalities be trusted to belong together?
We introduce \textbf{SHIFT-M3}, a lightweight text-based pre-fusion screen that
measures alignment-based consistency between two separately produced ECG text views:
an LLM-generated interpretation and a clinical report summary.
On 784{,}680 MEETI ECG records, SHIFT-M3 achieves
\textbf{\typeione\%} TPR@5\% FPR for full text-view swaps (AUROC \typeiauroc),
\textbf{\typeithree\%} for partial swaps (AUROC \typeiiiauroc), and
\textbf{\hardneg\%} for \mbox{label-matched} hard negatives
(AUROC \hardnegauroc) with only
573,569 parameters.
Compared with same-dataset lexical baselines, the gains are largest on partial
swaps and hard negatives, suggesting that the model is learning more than
surface overlap.
We also introduce the \textbf{CMST} (Conflict-type Multimodal Stress Test) evaluation taxonomy, a three-seed
stability study, a loss ablation, a temporal-tolerance sweep, and a
shared-token masking control.
The main remaining failure mode is longitudinal ambiguity: at the default
operating point, same-patient cross-visit pairs still produce \typeiifpr\%
Type-II false positives.
\end{abstract}

\section{Introduction}

Modern clinical ML systems increasingly fuse waveforms, reports, images, and
metadata to improve downstream prediction or triage
\citep{huang2020fusion,kline2022multimodal,zhang2020structuredunstructured,lyu2023multimodal,garriga2023mental}.
In practice, clinical information systems remain vulnerable to interoperability
and documentation failures that can associate one patient's data with another's
through clerical mistakes, workflow breakdowns, or software errors
\citep{hyvamaki2023hie,riplinger2020patient,grannis2019standardization,grannis2022referential}.
Standard benchmarks rarely test this failure mode: they assume the observed
multimodal pair is valid.
Yet a record-level linkage error can propagate into decision support, quality
dashboards, or retrospective research cohorts.
This problem sits adjacent to EHR data-quality assessment and patient matching,
but what remains less studied is \emph{content-level integrity checking} once a
multimodal record has already been assembled
\citep{weiskopf2013ehr,kahn2012framework,kahn2016harmonized,brown2013dataquality,lewis2023ehr,riplinger2020patient,li2022fellegi}.

Most multimodal representation learning methods optimize fusion, retrieval, or
alignment under the assumption that observed pairs are correct inputs
\citep{saporta2024symile,liu2024merl,liu2025kmerl,wang2025melp,wang2023finegrained,yan2023styleaware}.
This leaves three gaps for integrity detection.
First, \textbf{post-fusion detectors} are structurally disadvantaged once
inconsistent components have already been merged, because downstream predictors
are usually trained to map assembled records to outcomes rather than to audit
whether the assembly itself is valid.
Second, \textbf{identity-centric matching} depends on strong biometric signal;
recent work shows this can be useful for ECG images \citep{sangha2024biometric},
but in our ECG-text setting identity signal alone is too weak.
Third, many clinical workflows expose \textbf{text-rich but image-poor}
records, so approaches that require heavy image pipelines are operationally
brittle.

We therefore ask a different question: should the information have been fused at all?
Our hypothesis is that contrastively trained alignment between
\emph{separately produced views} of the same event is a stronger integrity
signal than latent identity matching.
This reframes the task from ``better multimodal prediction'' to ``pre-fusion
consistency estimation.''

We test this idea in the MEETI dataset \citep{meeti2026}, where each ECG record
contains two complementary text views:
(i) an \texttt{LLM\_Interpretation}, a detailed automated analysis, and
(ii) a short clinical \texttt{report}.
For a correctly linked record, both texts describe the same ECG.
For a swapped record, the interpretation describes Patient~A while the report
describes Patient~B, creating a detectable content mismatch.
SHIFT-M3 measures that mismatch directly with a lightweight pre-fusion model.
The resulting score is not a diagnosis; it is a triage signal that asks
whether the assembled record warrants integrity review.

Our contributions are as follows:
\begin{enumerate}[leftmargin=*,topsep=2pt,itemsep=1pt]
  \item A \textbf{problem formulation} of multimodal record integrity triage as
        pre-fusion alignment-based consistency estimation rather than standard
        multimodal prediction.
  \item A \textbf{lightweight method}, SHIFT-M3, that detects full and
        partial text-view swaps on 784K ECG records with only
        573,569 parameters.
  \item The \textbf{CMST taxonomy} (Conflict-type Multimodal Stress Test),
        which separates full text-view swaps, partial swaps, hard negatives, and
        longitudinal pairs.
  \item An \textbf{evaluation package} comprising
        same-dataset lexical baselines, frozen pretrained encoder baselines (MiniLM and Bio\_ClinicalBERT), three-seed stability analysis, a
        temporal sweep, a full operating-point frontier, and a shared-token masking control extended to all stressors.
\end{enumerate}

\subsection*{Generalizable Insights about Machine Learning in the Context of Healthcare}

A cross-patient linkage error is undetectable from either text view in isolation: each document is individually plausible, yet a downstream model processing the assembled pair would attribute findings from two different patients to one, potentially generating incorrect risk predictions or clinical alerts.
This study provides three generalizable insights for machine learning in
healthcare. First, multimodal clinical pipelines should treat record assembly as
an explicit source of risk rather than assuming that paired modalities are valid
inputs. Second, lightweight pre-fusion consistency screens can detect many
cross-component linkage failures before downstream fusion or prediction occurs,
even when each individual component appears clinically plausible. Third,
content-sensitive integrity screening creates a predictable tradeoff with
longitudinal tolerance: same-patient clinical change can resemble
cross-patient inconsistency unless identity-aware or temporal context is
incorporated. These findings suggest that integrity screening should be
evaluated as a distinct stage in multimodal healthcare pipelines, especially in
settings where silent record assembly errors can propagate into decision
support, quality measurement, or retrospective cohorts.

\section{Related Work}

Most multimodal clinical AI work is organized around a downstream prediction
task: combining images, EHR variables, waveforms, and text to improve
classification, risk stratification, or triage
\citep{huang2020fusion,kline2022multimodal}.
Concrete EHR systems follow the same pattern by fusing structured variables
with clinical notes for mortality, readmission, or crisis prediction
\citep{zhang2020structuredunstructured,lyu2023multimodal,garriga2023mental}.
Recent general multimodal representation-learning methods likewise treat
observed pairs as trustworthy supervision for fusion, retrieval, or alignment
\citep{saporta2024symile,chen2020simclr}.
These methods improve prediction, but they do not test whether the pair itself
is valid.

AI-enabled ECG research has expanded from single-task classification toward
broader representation learning and multimodal report-aware modeling
\citep{attia2021ai}.
Recent ECG-language methods align ECG signals with reports for zero-shot
classification, arbitrary-lead representation learning, or report generation
\citep{liu2024merl,wang2025melp,liu2025kmerl,wan2025meit}.
These systems show that ECGs and associated text contain rich shared clinical
content, which directly motivates our paired-view formulation.
However, they still presume that the interpretation and report presented during
training correspond to the same event.
Our goal is different: not to predict from paired modalities or generate text,
but to test whether two separately produced descriptions should be trusted as
belonging to the same record.

Adjacent medical vision-language work studies how to generate or verbalize
reports from correctly paired clinical inputs.
Recent papers emphasize fine-grained image-report alignment, low-label
self-training, style control, and longitudinal context for report generation
\citep{wang2023finegrained,wang2023selftraining,yan2023styleaware,dallaserra2023longitudinal}.
This literature treats cross-modal agreement as useful structure, but still
assumes the pair is trustworthy.
SHIFT-M3 targets the missing pre-fusion screening step.
Clinical informatics has separately developed strong frameworks for data
quality assessment and record linkage.
Foundational work formalizes data-quality dimensions for EHR reuse and
multisite research
\citep{weiskopf2013ehr,kahn2012framework,kahn2016harmonized,bian2020dataquality,brown2013dataquality,lewis2023ehr}.
Parallel patient-matching studies evaluate deterministic, probabilistic, and
machine learning approaches, including standardization-sensitive matching,
adaptive Fellegi-Sunter variants, and manual audits of real-world linkage
quality
\citep{riplinger2020patient,eggerth2019timeseries,grannis2019standardization,li2022fellegi,grannis2022referential,gupta2024manual}.
Patient-safety studies further show that interoperability and documentation
failures remain an active source of clinical risk \citep{hyvamaki2023hie}.
At the modality level, identity-centric contrastive learning can exploit
patient-specific biometric signal when the channel is strong enough, as shown
for ECG images \citep{sangha2024biometric}.

These literatures do not directly address the post-assembly question studied
here: once a multimodal record exists, can its internal content be screened for
cross-patient inconsistency?
SHIFT-M3 sits between multimodal representation learning and clinical
data-quality assessment. It leverages redundancy between two views of the same
event, but uses that redundancy for integrity screening rather than prediction.
Its distinguishing contribution is to recast multimodal consistency as a
\emph{pre-fusion integrity screening} problem and to evaluate it under explicit
record-level stress tests rather than standard downstream prediction metrics.

\section{Method}

\subsection{Problem Formulation}

Each record is represented as $r=(t_A,t_B)$, where $t_A$ is an LLM-derived ECG
interpretation and $t_B$ is the clinical report summary.
The task is to learn a score $s(t_A,t_B) \in [0,1]$ such that higher values
indicate likely cross-patient mismatch.
Screening operates at the level of a single patient event: the question is
whether the two text views assembled into one record describe the same
clinical event, not whether a patient's identity is consistent across
separate visits.

\begin{figure}[t]
\centering
\begin{minipage}[t]{0.49\textwidth}
  \centering
  \includegraphics[width=\linewidth]{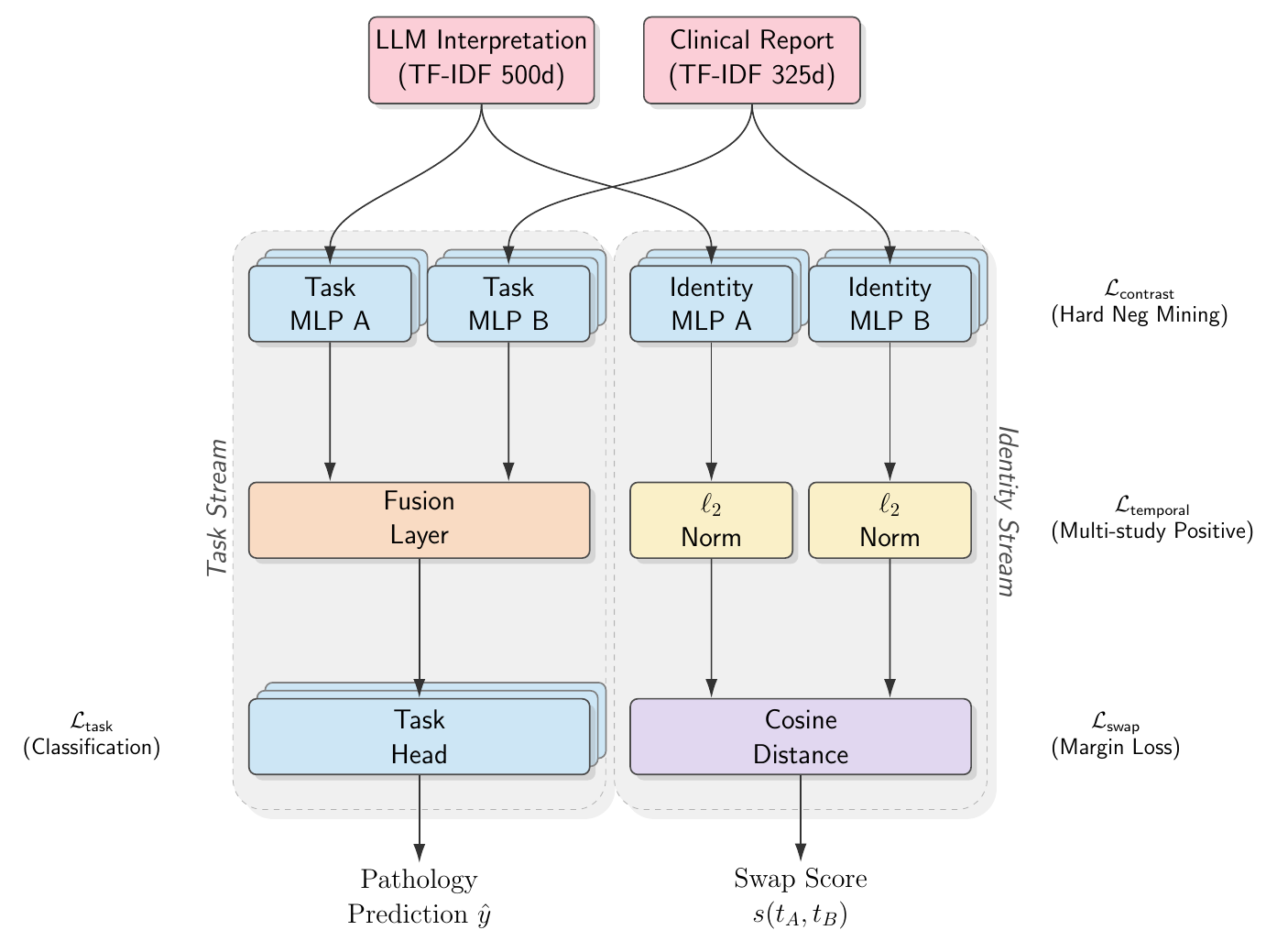}
\end{minipage}\hfill
\begin{minipage}[t]{0.49\textwidth}
  \centering
  \includegraphics[width=\linewidth]{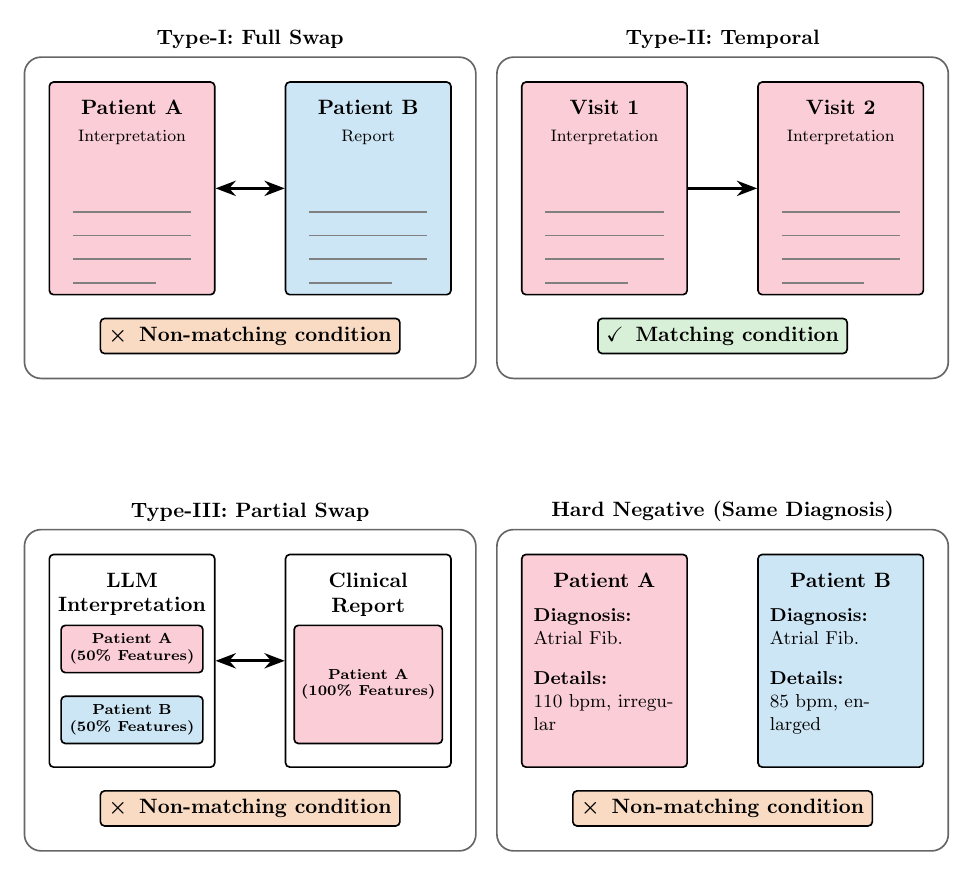}
\end{minipage}
\caption{Method overview. On the left, SHIFT-M3 encodes the interpretation and
report with separate modality-specific streams and produces a pre-fusion
conflict score before any downstream multimodal model is invoked. On the right, the
CMST taxonomy decomposes evaluation into full text-view swaps, partial swaps, hard
negatives, and same-patient temporal controls, making it possible to separate
cross-patient mismatch detection from longitudinal tolerance.}
\label{fig:method_miccai}
\end{figure}

We evaluate four stressors with the \textbf{CMST} taxonomy shown in
Figure~\ref{fig:method_miccai}:
\begin{enumerate}[leftmargin=*,topsep=2pt,itemsep=1pt]
  \item \textbf{Type-I (Full Text-view Swap):} replace $t_B$ with another
        patient's report, creating a complete mismatch between the two text
        views while leaving each view individually plausible.
  \item \textbf{Type-II (Temporal):} pair records from the same patient but
        different visits.
  \item \textbf{Type-III (Partial Swap):} replace half of the interpretation
        TF-IDF features in $t_A$ with features from another patient's
        interpretation, creating a mixed record in which only part of the
        evidence is inconsistent. This operation is performed in TF-IDF
        feature space and does not correspond to readable mixed text.
        Unlike real partial corruptions, which typically consist of
        coherent but misattributed readable text (such as copied-forward
        documentation or split-record merges), this substitution operates
        at the level of abstract feature overlap. Type-III is therefore a
        representation-space stress test that probes the model's
        sensitivity to partial feature-level mismatch.
  \item \textbf{Hard Negative:} pair with a different patient sharing the same
        weak abnormal/normal label, where that auxiliary label is derived by
        keyword matching over the report text rather than by expert
        adjudication.
\end{enumerate}
Types I, III, and Hard Negative are failures to detect; Type II is a
legitimate longitudinal pair that should \emph{not} be flagged.
All four stressors are constructed algorithmically from
clean MEETI records. No documented real-world linkage errors are included
in the evaluation, and the degree to which these synthetic stressors
reflect actual EHR assembly failures remains an open empirical question.

\subsection{SHIFT-M3 Architecture}

SHIFT-M3 is a jointly trained dual-stream text encoder, analogous in design to BGE-M3~\citep{chen2025bge}, with separate TF-IDF vocabularies for interpretations and reports (500 and 325 features, respectively).
The vocabulary sizes are motivated by a coverage analysis on the training split: clinical reports (5--20 tokens per document) reach 99.9\% TF-IDF coverage at 325 features, and LLM interpretations reach 75.1\% coverage at 500 features.
To verify sufficiency, we trained SHIFT-M3 with LLM vocabulary size $\in \{500, 1000, 1500\}$: Type-I AUROC was 0.996 in all cases and TPR@5\% varied by less than 0.06~pp, confirming that 500 features captures the full discriminative vocabulary.
An identity stream produces normalized embeddings
$z_A^{id}, z_B^{id}$ and defines a conflict score via cosine distance:
\begin{equation}
s(t_A,t_B) = \frac{1 - \cos(z_A^{id}, z_B^{id})}{2}.
\end{equation}

SHIFT-M3 contains separate auxiliary task and identity streams. The
conflict score is computed exclusively from the identity stream.
Each MLP follows a Linear($d_{in}$, 256) $\to$ ReLU $\to$ BatchNorm $\to$
Dropout(0.2) $\to$ Linear(256, $d_{out}$) design, where the identity
stream maps to a 128-dimensional embedding ($d_{out}{=}128$).
The evaluated model has exactly 573,569 trainable parameters.

\subsection{Training Objective}

{Training combines an auxiliary classification loss, a contrastive loss,
a temporal positive loss, and an explicit swap margin loss:}
\begin{equation}
\mathcal{L} = \mathcal{L}_{task} + \lambda_{ctr}\mathcal{L}_{ctr}
            + \lambda_{temp}\mathcal{L}_{temp} + \mathcal{L}_{swap}.
\end{equation}
For a batch of size $B$, let $d_{ij}=1-\cos(z_{A,i}^{id},z_{B,j}^{id})$ and
$[x]_+=\max(x,0)$.
The contrastive loss uses this unscaled cosine distance, whereas reported
conflict scores use the scaled score $s$ in Eq.~(1).
The task loss is binary cross-entropy between the auxiliary task-head
output and the weak abnormal/normal label.
The contrastive term uses diagonal pairs as positives and the closest
off-diagonal report in the batch as the mined negative:
\begin{equation}
\mathcal{L}_{ctr} =
\frac{1}{B}\sum_i d_{ii}^{2}
+\frac{1}{B}\sum_i \left[m-\min_{j\neq i} d_{ij}\right]_+^{2}.
\end{equation}
The temporal term treats same-subject, different-visit pairs as positives and
penalizes high conflict scores,
$\mathcal{L}_{temp}=K^{-1}\sum_{k=1}^{K}s(t_{A,i_k},t_{B,j_k})$, where
$i_k$ and $j_k$ share a subject identifier but correspond to different ECG
records. If a batch contains fewer than two eligible temporal pairs, this term
is set to zero. Finally, the swap term applies an in-batch cyclic report
permutation $\pi$ and pushes swapped pairs above the margin:
\begin{equation}
\mathcal{L}_{swap}=\frac{1}{B}\sum_i
\left[m-s(t_{A,i},t_{B,\pi(i)})\right]_+ .
\end{equation}
We set $\lambda_{ctr} = 1.0$, $\lambda_{temp} = 0.5$, and swap margin $m = 0.5$
as the default configuration.
The contrastive term brings matched pairs together while pushing hard negatives
apart; the temporal term reduces penalties on same-patient cross-visit pairs;
and the swap term explicitly pushes synthetic swaps toward higher conflict
scores.

\subsection{Dataset and Setup}

MEETI is a publicly available dataset of 784{,}680 ECG records derived from
MIMIC-IV-ECG \citep{meeti2026,johnson2023mimic}; we use it as released without
modification.
The \texttt{LLM\_Interpretation} field is generated by GPT-4o from extracted ECG parameters together with the associated clinical report~\citep{meeti2026}; the \texttt{report} field is the independently released clinical report text from the source ECG record.
We therefore treat the two fields as separately produced textual views of the
same ECG event, but not as independent expert-adjudicated diagnoses.
Because both views share a common derivation from the clinical report, perfect cross-view independence cannot be assumed.
For cross-patient swap stressors, Patient~A's interpretation is paired with Patient~B's independent record, breaking paired-view consistency; post-masking Type-I AUROC of \maskedtypeiauroc\ confirms that the alignment signal reflects clinical content beyond shared stylistic overlap.
We acknowledge this derivation dependency as a limitation and note it in Section~\ref{sec:discussion}.
We use a patient-level 80/10/10 split to prevent subject leakage (Table~\ref{tab:dataset_miccai}).
Training uses Adam, learning rate $3 \times 10^{-4}$, batch size 128, 10
epochs, and 128-dimensional embeddings.

\begin{table}[htbp]
\centering
\caption{Patient-level split in the MEETI dataset pipeline. Splitting by subject
prevents leakage of repeated visits across train, validation, and test.}
\label{tab:dataset_miccai}
\small
\setlength{\tabcolsep}{5pt}
\begin{tabular}{lrrr}
\toprule
\textbf{Split} & \textbf{Records} & \textbf{Subjects} & \textbf{Multi-Visit} \\
\midrule
Train      & 629,386 & 128,477 & 82,414 \\
Validation &  76,742 &  16,059 & 10,342 \\
Test       &  78,552 &  16,061 & 10,245 \\
\midrule
\textbf{Total} & \textbf{784,680} & \textbf{160,597} & \textbf{103,001} \\
\bottomrule
\end{tabular}
\end{table}
\section{Results}
\label{sec:results}

\subsection{Main Performance and Same-dataset Baselines}

Table~\ref{tab:main_miccai} compares SHIFT-M3 with same-dataset lexical
baselines under a common 5\% clean-pair false-positive budget.
The lexical baselines are as follows: \emph{word cosine} and \emph{char cosine} compute cosine similarity over TF-IDF vectors with no learned parameters; \emph{LogReg} fits approximately ten hand-crafted pairwise similarity features using fewer than 50 weights.
SHIFT-M3 reaches \typeione\% Type-I TPR, \typeithree\% Type-III TPR, and
\hardneg\% hard-negative TPR, outperforming all lexical baselines on every
cross-patient stressor.
The strongest baseline is a logistic regression over hand-crafted similarity
features, but it still trails substantially on both full text-view and partial
swaps.
This difference matters because the partial-swap setting is the closest proxy
to subtle record corruption: simple overlap can detect many obvious complete
text-view swaps,
yet it degrades sharply once only part of the evidence is inconsistent.
For example, word TF-IDF cosine reaches Type-I AUROC \wordtypeiauroc, but only
Type-III AUROC \wordtypeiiiauroc, whereas SHIFT-M3 maintains
Type-III AUROC \typeiiiauroc.
The temporal column shows a different ordering: the lexical baselines achieve
substantially lower Type-II FPR because they are less sensitive to clinically
meaningful semantic drift across visits.
The empirical picture is therefore not that SHIFT-M3 is uniformly better, but
that it trades temporal specificity for much stronger sensitivity to
cross-patient corruption.

\begin{table}[htbp]
\centering
\caption{Frozen pretrained encoder baselines versus SHIFT-M3 on the held-out test set. Both encoders score pairs via cosine distance between independently encoded embeddings without fine-tuning. Lower Type-II FPR in frozen models reflects weak conflict sensitivity across all pairs, not better longitudinal tolerance. Best value per column in bold; higher is better for AUROC, lower is better for Type-II FPR.\label{tab:frozen_baselines}}

\small
\setlength{\tabcolsep}{4pt}
\begin{tabular}{lcccc}
\toprule
\textbf{Model} &
\shortstack{\textbf{Type-I}\\\textbf{AUROC}} &
\shortstack{\textbf{Type-III}\\\textbf{AUROC}} &
\shortstack{\textbf{Hard Neg}\\\textbf{AUROC}} &
\shortstack{\textbf{Type-II}\\\textbf{FPR@5\%}} \\
\midrule
MiniLM (frozen) & \minilmtypeiauroc & \minilmtypeiiiauroc & \minilmhardnegauroc & \minilmtypeiifpr\% \\
Bio\_ClinicalBERT (frozen) & \biocberttypeiauroc & \biocberttypeiiiauroc & \biocberthardnegauroc & \textbf{\biocberttypeiifpr\%} \\
\midrule
\textbf{SHIFT-M3} & \textbf{\typeiauroc} & \textbf{\typeiiiauroc} & \textbf{\hardnegauroc} & \typeiifpr\% \\
\bottomrule
\end{tabular}
\end{table}

\paragraph*{Comparison with frozen pretrained encoders:}
To assess whether supervised contrastive training adds value beyond off-the-shelf text representations, we evaluated two frozen contextual encoders as additional baselines: \textsc{MiniLM}~\citep{wang2020minilm} and \textsc{Bio\_ClinicalBERT}~\citep{alsentzer2019clinicalbert}, each scoring pairs via cosine distance between independently encoded LLM interpretation and report embeddings without any fine-tuning.
Table~\ref{tab:frozen_baselines} reports results on the three cross-patient stressors.
Both frozen encoders fall substantially below SHIFT-M3 across all stressors, with \textsc{Bio\_ClinicalBERT} performing near chance on Type-III and Hard Negatives.
Their lower Type-II FPR is not evidence of better longitudinal tolerance: it reflects uniformly weak conflict sensitivity across all pair types.
These results demonstrate that task-specific contrastive training over TF-IDF features is a more effective strategy for record integrity screening than frozen clinical embedding cosine similarity.

\begin{table}[htbp]
\centering
\caption{Held-out CMST performance. SHIFT-M3 improves substantially over
same-dataset lexical baselines on full text-view swaps, partial swaps, and hard
negatives. Type-II is a false-positive rate, so lower is better. Best value per
column in bold.}
\label{tab:main_miccai}
\scriptsize
\begin{minipage}[htbp]{0.52\textwidth}
\centering
\textbf{(a) Recall at 5\% clean-pair FPR}
\vspace{0.3em}

\setlength{\tabcolsep}{3pt}
\begin{tabular}{lccc}
\toprule
\textbf{Model} & \textbf{T-I} & \textbf{T-III} & \textbf{HN} \\
\midrule
Word cosine & \wordtypeione & \wordtypeithree & \wordhardneg \\
Char cosine & \chartypeione & \chartypeithree & \charhardneg \\
LogReg & \logregtypeione & \logregtypeithree & \logreghardneg \\
\midrule
\textbf{SHIFT-M3} & \textbf{\typeione} & \textbf{\typeithree} & \textbf{\hardneg} \\
\bottomrule
\end{tabular}
\end{minipage}\hfill
\begin{minipage}[htbp]{0.48\textwidth}
\centering
\textbf{(b) Ranking quality and temporal error}
\vspace{0.3em}

\setlength{\tabcolsep}{2.8pt}
\begin{tabular}{lcccc}
\toprule
\textbf{Model} &
\shortstack{\textbf{T-I}\\\textbf{AUC}} &
\shortstack{\textbf{T-III}\\\textbf{AUC}} &
\shortstack{\textbf{HN}\\\textbf{AUC}} &
\shortstack{\textbf{T-II}\\\textbf{FPR}} \\
\midrule
Word cosine & \wordtypeiauroc & \wordtypeiiiauroc & \wordhardnegauroc & \wordtypeiifpr \\
Char cosine & \chartypeiauroc & \chartypeiiiauroc & \charhardnegauroc & \textbf{\chartypeiifpr} \\
LogReg & \logregtypeiauroc & \logregtypeiiiauroc & \logreghardnegauroc & \logregtypeiifpr \\
\midrule
\textbf{SHIFT-M3} & \textbf{\typeiauroc} & \textbf{\typeiiiauroc} & \textbf{\hardnegauroc} & \typeiifpr \\
\bottomrule
\end{tabular}
\end{minipage}
\end{table}

\begin{figure}[H]
\centering
\begin{minipage}[t]{0.48\textwidth}
\centering
\includegraphics[width=0.96\linewidth]{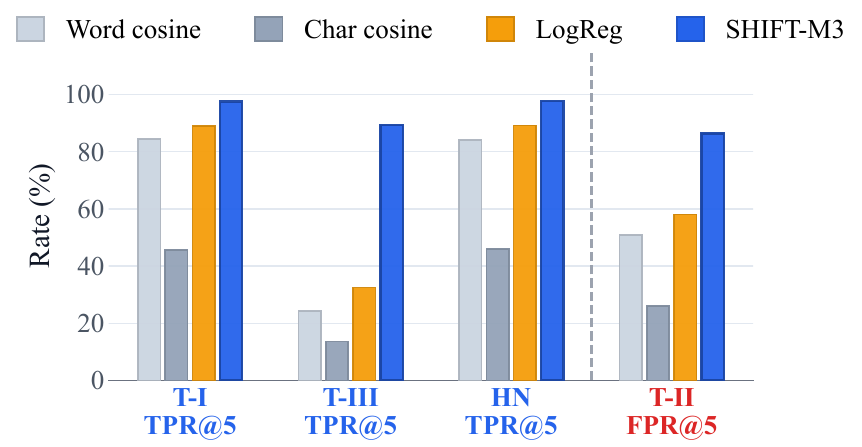}
\captionof{figure}{Performance at the 5\% clean-pair FPR budget.
SHIFT-M3 outperforms all lexical baselines on every cross-patient stressor,
with the largest margin on partial swaps, whereas same-patient longitudinal
pairs remain the dominant residual error source.}
\label{fig:results_miccai}
\end{minipage}\hfill
\begin{minipage}[t]{0.44\textwidth}
\centering
\includegraphics[width=0.96\linewidth]{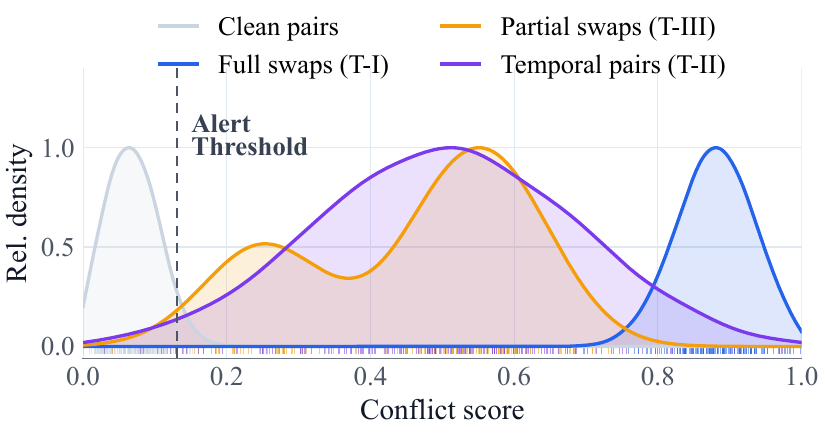}
\captionof{figure}{Conflict-score densities by pair type.
Clean pairs and full swaps are well separated, whereas partial swaps and
temporal pairs occupy the intermediate region near the alert threshold.}
\label{fig:score_distributions_miccai}
\end{minipage}
\end{figure}

Figures~\ref{fig:results_miccai} and~\ref{fig:score_distributions_miccai}
complement Table~\ref{tab:main_miccai} with a visual breakdown of the
per-stressor margins and the score distributions that underlie them.
The gap is largest on partial swaps, where lexical overlap degrades sharply
but SHIFT-M3 remains consistent across seeds.

\begin{figure}[H]
\centering
\includegraphics[width=0.95\textwidth]{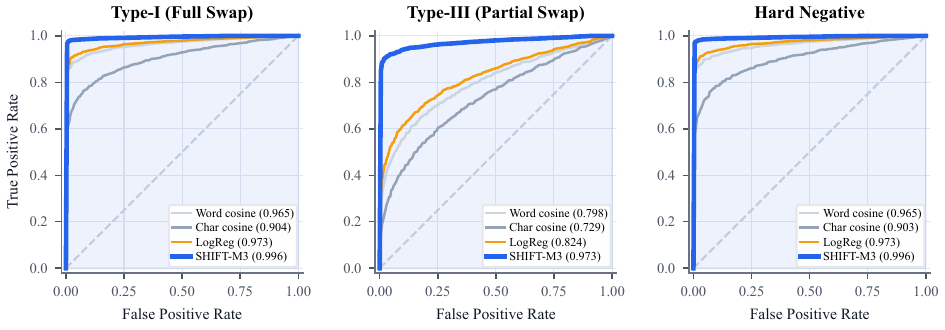}
\caption{ROC curves across the three cross-patient stressors.
SHIFT-M3 (blue) consistently hugs the upper-left corner,
with the largest margin on partial swaps (center panel, AUROC \typeiiiauroc\ vs.\ 0.824).
Each curve is annotated with its AUROC. The dashed diagonal represents random chance.}
\label{fig:roc_miccai}
\end{figure}

Figure~\ref{fig:roc_miccai} confirms these rankings across the full operating
range.
SHIFT-M3 achieves the highest AUROC on all three cross-patient stressors, with
the widest margin on partial swaps (\typeiiiauroc\ vs.\ 0.824 for the best
lexical baseline).

\subsection{Stability Across Seeds}

Table~\ref{tab:e28_perseed_miccai} reports all three completed seeds, and
Table~\ref{tab:e28_miccai} summarizes the mean and standard deviation.
Variance is minimal for the main swap-detection metrics: Type-I AUROC remains
within $10^{-4}$ and Type-III AUROC within roughly $2 \times 10^{-3}$ across
runs.
The larger spread in Type-II FPR is informative rather than alarming, because
longitudinal pairs sit near the conceptual boundary between legitimate change
and suspicious inconsistency.

\begin{table}[htbp]
\centering
\caption{Per-seed held-out metrics from the three-run stability analysis. The main detection metrics vary
little across seeds, indicating that the reported gains are not artifacts of a
single run.}
\label{tab:e28_perseed_miccai}
\scriptsize
\setlength{\tabcolsep}{3pt}
\resizebox{\textwidth}{!}{%
\begin{tabular}{cccccccc}
\toprule
\textbf{Seed} & \textbf{Type-I TPR@5} & \textbf{Type-III TPR@5} & \textbf{Hard Neg TPR@5} & \textbf{Type-II FPR@5} & \textbf{Type-I AUROC} & \textbf{Type-III AUROC} & \textbf{Hard Neg AUROC} \\
\midrule
42 & 97.55 & 89.40 & 97.83 & 86.42 & 0.9962 & 0.9725 & 0.9964 \\
43 & 97.66 & 91.41 & 97.81 & 87.86 & 0.9963 & 0.9770 & 0.9961 \\
44 & 97.47 & 90.02 & 97.57 & 86.56 & 0.9961 & 0.9737 & 0.9960 \\
\bottomrule
\end{tabular}
}
\end{table}

\begin{table}[htbp]
\centering
\caption{Three-seed stability summary. Small standard deviations on
Type-I, Type-III, and hard-negative metrics indicate stable optimization.}
\label{tab:e28_miccai}
\resizebox{0.325\textwidth}{!}{%
\setlength{\tabcolsep}{4pt}
\begin{tabular}{lcc}
\toprule
\textbf{Metric} & \textbf{Mean} & \textbf{Std} \\
\midrule
Type-I TPR@5 & 97.56 & 0.09 \\
Type-III TPR@5 & 90.28 & 1.03 \\
Hard Neg TPR@5 & 97.73 & 0.15 \\
Type-II FPR@5 & 86.95 & 0.79 \\
Type-I AUROC & 0.9962 & 0.0001 \\
Type-III AUROC & 0.9744 & 0.0023 \\
Hard Neg AUROC & 0.9962 & 0.0002 \\
\bottomrule
\end{tabular}
}
\end{table}

\subsection{Loss Ablation}

\begin{figure}[t]
\centering
\begin{minipage}[t]{0.47\textwidth}
\centering
\includegraphics[width=\linewidth]{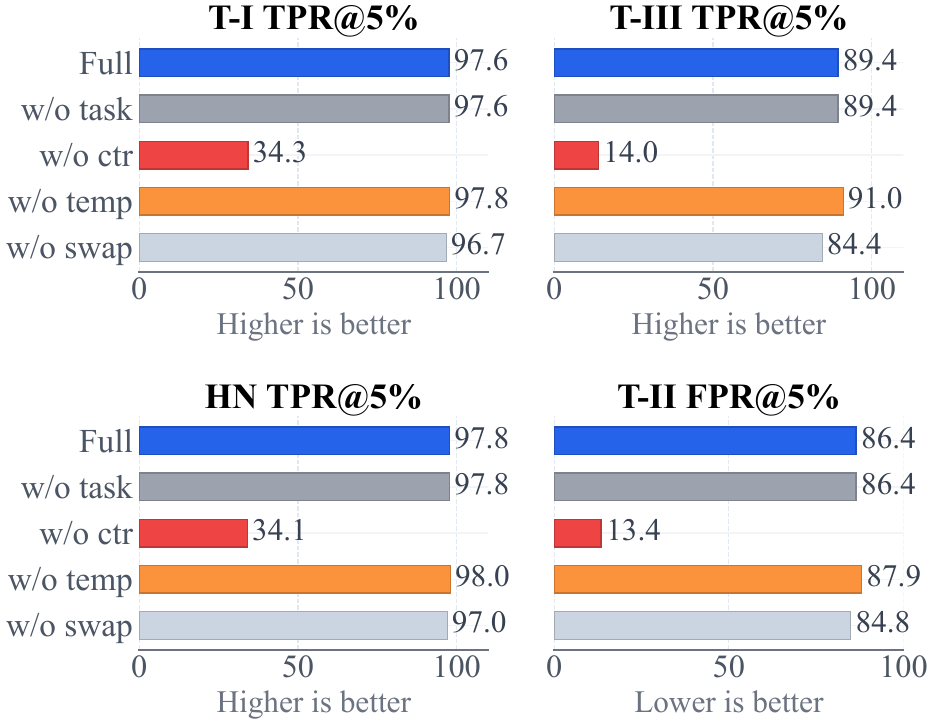}
\captionof{figure}{Loss ablation across mismatch conditions. The contrastive term provides
the primary detection signal, whereas the swap and temporal terms specifically
tune partial-swap sensitivity and longitudinal tolerance.}
\label{fig:e29_ablation_miccai}
\end{minipage}\hfill
\begin{minipage}[t]{0.47\textwidth}
\centering
\includegraphics[width=\linewidth]{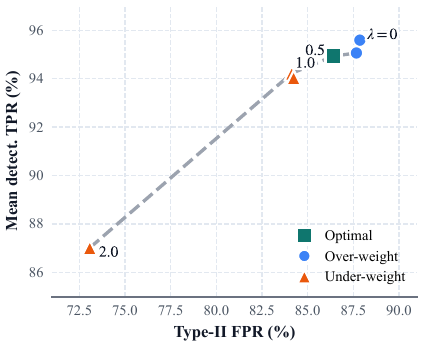}
\captionof{figure}{Temporal-weight sweep. The tradeoff frontier demonstrates
that moderate temporal weighting ($\lambda_{temp}=0.5$) optimally balances
same-patient longitudinal tolerance against cross-patient detection.}
\label{fig:e30_tradeoff_miccai}
\end{minipage}
\end{figure}

Table~\ref{tab:e29_miccai} and Figure~\ref{fig:e29_ablation_miccai} report
the full ablation results.
Removing the contrastive term causes a pronounced collapse across all swap
types, showing that aligned representation learning is the primary mechanism
behind the detector.
Removing the explicit swap loss also degrades performance, especially on
partial swaps, which indicates that synthetic conflict supervision is not
redundant with the contrastive objective alone.
By contrast, removing the temporal term slightly improves raw detection but
worsens Type-II false positives, confirming that its role is to trade some
sensitivity for better longitudinal tolerance.
Setting the auxiliary task-loss weight to zero produced identical
conflict-detection metrics, as expected because the task and identity streams
have disjoint parameters. Accordingly, the reported conflict scores are
optimized through the contrastive, temporal, and swap objectives. The reported
results retain the evaluated 573,569-parameter architecture.

\begin{table}[htbp]
\centering
\caption{Loss ablation. The contrastive and swap losses provide most
of the integrity signal, whereas the temporal term mainly moderates false
positives on same-patient longitudinal pairs. The task-loss variant matches
Full because the task and identity streams share no parameters.}
\label{tab:e29_miccai}
\scriptsize
\setlength{\tabcolsep}{3pt}
\resizebox{\textwidth}{!}{%
\begin{tabular}{lccccccc}
\toprule
\textbf{Variant} & \textbf{Type-I TPR@5} & \textbf{Type-III TPR@5} & \textbf{Hard Neg TPR@5} & \textbf{Type-II FPR@5} & \textbf{Type-I AUROC} & \textbf{Type-III AUROC} & \textbf{Hard Neg AUROC} \\
\midrule
Full & 97.55 & 89.40 & 97.83 & 86.42 & 0.9962 & 0.9725 & 0.9964 \\
No task loss & 97.55 & 89.40 & 97.83 & 86.42 & 0.9962 & 0.9725 & 0.9964 \\
No contrastive loss & 34.28 & 14.01 & 34.12 & 13.37 & 0.8445 & 0.6634 & 0.8443 \\
No temporal loss & 97.76 & 91.03 & 97.98 & 87.86 & 0.9964 & 0.9760 & 0.9967 \\
No swap loss & 96.73 & 84.40 & 97.04 & 84.79 & 0.9945 & 0.9575 & 0.9949 \\
\bottomrule
\end{tabular}
}
\end{table}

\subsection{Temporal Tolerance Sweep}

The temporal-weight sweep exposes a real tradeoff rather than a trivial tuning oversight.
Increasing the temporal weight improves same-patient tolerance, but too much
temporal encouragement suppresses true swap sensitivity.
Table~\ref{tab:e30_miccai} lists all completed sweep runs and
Figure~\ref{fig:e30_tradeoff_miccai} plots the resulting tradeoff frontier;
together they show a clear pattern: the highest raw detection appears when the temporal loss is removed,
whereas aggressive temporal weighting lowers Type-II FPR at a substantial cost
to Type-I and Type-III recall.
The most defensible operating point in the current grid is therefore the
compromise setting $\lambda_{temp}=\sweeplambda$ with swap margin
\sweepmargin, which reduces Type-II FPR relative to the default without
eroding detection as sharply as the aggressive setting.
We recommend the default setting ($\lambda_{temp}=0.5$) for first-linkage screening contexts, where cross-patient swap detection is the primary objective, and the compromise setting ($\lambda_{temp}=1.0$) for high-revisit cardiology departments, where repeated same-patient ECGs are common and reducing alert burden is a practical priority.

\paragraph*{Full operating-point frontier:}
To clarify whether any threshold setting can resolve longitudinal ambiguity, Table~\ref{tab:frontier_miccai} reports the full operating-point frontier at clean-pair FPR budgets from 1\% to 20\%.
Type-II FPR ranges from \frontierOneTII\% at a 1\% budget to \frontierTwentyTII\% at 20\%, while Type-I TPR remains above 96\% throughout.
No threshold choice resolves the longitudinal false-positive problem: tightening the threshold reduces both cross-patient recall and same-patient errors simultaneously, because genuine longitudinal change and cross-patient swaps produce similar text divergence patterns when only text content is available.

\begin{table}[htbp]
\centering
\caption{Operating-point frontier across clean-pair FPR budgets. Type-III results use the text-splice variant. No threshold setting resolves longitudinal ambiguity: Type-II FPR remains elevated across the full range.}
\label{tab:frontier_miccai}
\resizebox{0.6\textwidth}{!}{%
\begin{tabular}{ccccc}
\toprule
\shortstack{\textbf{Clean-pair}\\\textbf{FPR budget}} &
\shortstack{\textbf{Type-I}\\\textbf{TPR@thr}} &
\shortstack{\textbf{Type-III}\\\textbf{TPR@thr}} &
\shortstack{\textbf{Hard Neg}\\\textbf{TPR@thr}} &
\shortstack{\textbf{Type-II}\\\textbf{FPR@thr}} \\
\midrule
1\%  & \frontierOneTI\%  & \frontierOneTIII\%  & \frontierOneHN\%  & \frontierOneTII\%  \\
2\%  & 97.0\% & 69.0\% & 97.3\% & 80.2\% \\
5\%  & \frontierFiveTI\%  & \frontierFiveTIII\%  & \frontierFiveHN\%  & \frontierFiveTII\%  \\
10\% & \frontierTenTI\%  & \frontierTenTIII\%  & \frontierTenHN\%  & \frontierTenTII\%  \\
20\% & \frontierTwentyTI\% & \frontierTwentyTIII\% & \frontierTwentyHN\% & \frontierTwentyTII\% \\
\bottomrule
\end{tabular}
}
\end{table}

\begin{table}[H]
\centering
\caption{Completed temporal-weight sweep runs. Larger temporal weight lowers
Type-II false positives, but the gain comes at the cost of weaker cross-patient
detection.}
\label{tab:e30_miccai}
\scriptsize
\setlength{\tabcolsep}{3.5pt}
\resizebox{0.9\textwidth}{!}{%
\begin{tabular}{lcccccccc}
\toprule
\textbf{Config} & \boldmath$\lambda_{temp}$ & \textbf{Margin} &
\shortstack{\textbf{Type-I}\\\textbf{TPR@5}} &
\shortstack{\textbf{Type-III}\\\textbf{TPR@5}} &
\shortstack{\textbf{Hard Neg}\\\textbf{TPR@5}} &
\shortstack{\textbf{Type-II}\\\textbf{FPR@5}} &
\shortstack{\textbf{Type-I}\\\textbf{AUROC}} &
\shortstack{\textbf{Type-III}\\\textbf{AUROC}} \\
\midrule
No temporal & 0.0 & 0.5 & 97.76 & 91.03 & 97.98 & 87.86 & 0.9964 & 0.9760 \\
Low temporal & 0.25 & 0.5 & 97.66 & 90.64 & 97.89 & 87.67 & 0.9963 & 0.9747 \\
Default & 0.5 & 0.5 & \typeione & \typeithree & \hardneg & \typeiifpr & \typeiauroc & \typeiiiauroc \\
Compromise & \sweeplambda & \sweepmargin & \sweeptypeione & \sweeptypeithree & \sweephardneg & \sweeptypeiifpr & \sweeptypeiaurocfull & \sweeptypeiiiaurocfull \\
Margin 0.6 & 1.0 & 0.6 & 97.39 & 88.62 & 97.58 & 84.23 & 0.9957 & 0.9693 \\
Aggressive & \agglambda & \aggmargin & \aggtypeione & \aggtypeithree & \agghardneg & \aggtypeiifpr & \aggtypeiauroc & \aggtypeiiiauroc \\
\bottomrule
\end{tabular}
}
\end{table}

\subsection{Shortcut Control and Audit Export}
\label{sec:shortcut}

The shared-token masking control tests whether the model succeeds mainly because the two text
views share obvious tokens.
Masking shared tokens causes a substantial but not catastrophic drop, which
shows that lexical overlap contributes signal but does not fully explain the
result.
We extend the masking analysis beyond Type-I to cover the text-splice Type-III variant and Hard Negatives, addressing a key reviewer request.
For the text-splice Type-III stressor (source first half $+$ donor second half), performance drops substantially under masking (AUROC $\textspliceauroc \to \textsplicemaskedauroc$; TPR@5\% $\textsplicetprfive\% \to \textsplicetprfivemasked\%$), confirming that partial-swap detection is more lexically dependent than full-swap detection.
In contrast, Hard Negative detection remains strong post-masking (AUROC $\hardnegauroc \to \maskedhardnegauroc$), indicating that the model learns beyond surface overlap for the label-matched case.
The table caption distinguishes the text-splice Type-III variant used here from the feature-space substitution reported in Table~\ref{tab:main_miccai}, whose AUROC of \typeiiiauroc\ should not be directly compared.
Table~\ref{tab:e31_pair_miccai} summarizes the masking control metrics and
the audit table sizes.
To support later qualitative analysis, we also exported three audit tables of
250 examples each at the main 5\% clean-pair threshold.

\begin{table}[htbp]
\centering
\caption{Compact control summaries shown side by side. Left: shared-token masking
control. Right: audit tables exported at the main 5\% clean-pair threshold.}
\label{tab:e31_pair_miccai}
\scriptsize
\begin{minipage}[htbp]{0.58\textwidth}
\centering
\textbf{(a) Shared-token masking control}
\vspace{0.3em}

\setlength{\tabcolsep}{3pt}
\resizebox{\linewidth}{!}{%
\begin{tabular}{lcccc}
\toprule
\textbf{Setting} & \textbf{Thr.} & \textbf{T-I TPR@5} & \textbf{T-I AUC} & \textbf{Clean mean} \\
\midrule
Original & 0.131 & \typeione & \typeiauroc & 0.067 \\
Masked shared tokens & 0.452 & \maskedtypeione & \maskedtypeiauroc & 0.212 \\
\bottomrule
\end{tabular}
}
\end{minipage}\hfill
\begin{minipage}[htbp]{0.37\textwidth}
\centering
\textbf{(b) Audit table sizes at the 5\% FPR threshold}
\vspace{0.3em}

\setlength{\tabcolsep}{4pt}
\begin{tabular}{lc}
\toprule
\textbf{Table} & \textbf{Rows} \\
\midrule
Clean FP & 250 \\
Type-I FN & 250 \\
Type-II FP & 250 \\
\bottomrule
\end{tabular}
\end{minipage}
\end{table}

\section{Qualitative Error Analysis}
\label{sec:qualitative}

To understand the practical failure modes of our text-based alignment approach, we
extracted the top-scoring false positives (clean records flagged as swaps) and
lowest-scoring false negatives (swaps flagged as clean) at the 5\% FPR threshold.
Table~\ref{tab:qualitative} illustrates two representative edge cases.

In the False Positive case (Table~\ref{tab:qualitative}a), the LLM generates a
granular, step-by-step reasoning trace that notes minor deviations (\emph{bradycardia}, \emph{fluctuations}) before ultimately concluding the ECG is normal. The clinical report, however, skips the intermediate noise and directly asserts a \emph{Normal ECG}. The model penalizes this difference in granularity, incorrectly flagging the clean pair as a content mismatch.

Conversely, the False Negative case (Table~\ref{tab:qualitative}b) highlights a fundamental vulnerability of any strictly text-based sentinel. Here, the report from Case B was paired with the interpretation from Case A. Because both cases had similar findings (\emph{Sinus rhythm with PVCs}), the two text views were nearly identical. The text-text contrastive model successfully recognized the lexical equivalence between the views, but in doing so, failed to detect the underlying cross-patient mismatch. Resolving such clinically similar collisions requires independent verification of identity attributes such as demographics, which is outside the scope of the present text-only model.

\begin{table}[htbp]
\centering
\caption{Representative false positive and false negative cases at the 5\% FPR threshold. Terms contributing to model conflict are highlighted in \textcolor{red}{red}; terms consistent across both views are highlighted in \textcolor{teal}{teal}.}
\label{tab:qualitative}
\scriptsize
\setlength{\tabcolsep}{6pt}
\renewcommand{\arraystretch}{1.3}
\begin{tabular}{@{}lp{0.82\textwidth}@{}}
\toprule
\multicolumn{2}{c}{\textbf{(a) False Positive: Clean Pair Flagged as Swap}} \\
\midrule
\textbf{LLM Interp.} & The ECG image analysis begins with the global features revealing a \textcolor{red}{heart rate of 40 bpm}, which suggests \textcolor{red}{bradycardia}... \textcolor{red}{PR interval measurements vary significantly}... \textcolor{red}{large fluctuations in P and QRS amplitudes}... conclusion stands with a sinus rhythm \textcolor{teal}{without prominent abnormalities}, aligning with the representation of a \textcolor{teal}{normal ECG}. \\
\textbf{Clin. Report} & Sinus rhythm, \textcolor{teal}{Normal ECG} \\
\midrule
\multicolumn{2}{c}{\textbf{(b) False Negative: Full Text-view Swap Flagged as Clean Match}} \\
\midrule
\textbf{Case A LLM} & ...overall heart rate is 81 bpm, suggesting a \textcolor{teal}{normal sinus rhythm}. However... indicates the presence of \textcolor{red}{premature ventricular contractions (PVCs)}, aligning with the finding of a \textcolor{red}{borderline ECG}. \\
\textbf{Case B Report} & \textcolor{teal}{Sinus rhythm} with \textcolor{red}{PVC(s)}, \textcolor{red}{Borderline ECG} \\
\bottomrule
\end{tabular}
\end{table}

\section{Discussion}

\label{sec:discussion}
The central finding is that, in this setting, record integrity is better
captured by contrastively trained alignment between \emph{separately produced views} of
the same event than by generic lexical similarity or latent identity matching
alone.
A compact text-text model with task-specific contrastive training substantially outperforms both same-dataset lexical baselines and frozen pretrained encoders on the harder stressors despite using simpler features.
When the detailed LLM interpretation and short clinical report diverge
substantially, that disagreement is often a stronger integrity cue than a
single global similarity score.
This is consistent with recent ECG-language and medical report-generation work,
which shows that paired clinical views encode dense shared semantics beyond
surface token overlap \citep{liu2024merl,wang2025melp,wang2023finegrained,yan2023styleaware}.
The main remaining tradeoff is between content sensitivity and temporal
tolerance.

The Type-II FPR of \typeiifpr\% is not merely a tuning issue.
In a typical cardiology department where a substantial fraction of ECG records
are longitudinal re-visits, this rate would generate a large alert burden
(e.g., roughly 864 false alerts per 1{,}000 same-patient cross-visit pairs
screened), motivating the staged deployment model described below.
A patient can be stable in identity yet genuinely inconsistent in content
across visits because new ischemia, conduction changes, or treatment effects
alter the ECG narrative.
Without an explicit identity variable, SHIFT-M3 interprets strong content
drift as suspicious.
This is also why the lexical baselines appear better on Type-II: they are
less responsive to semantic change overall, which helps temporal tolerance but
hurts corruption detection on the harder cross-patient stressors.
SHIFT-M3 is therefore better suited to first-pass screening than final
adjudication for longitudinal records.

The temporal sweep moderates this tension only partially, and the loss ablation
supports the same interpretation from another angle: contrastive alignment and
explicit swap supervision carry most of the integrity signal, whereas the
temporal term mainly controls how aggressively the model treats legitimate
clinical change.
That limitation echoes a broader theme in longitudinal clinical modeling:
temporal context is informative, but it cannot be treated as noise when the
underlying patient state is genuinely evolving \citep{dallaserra2023longitudinal,attia2021ai}.

The practical deployment model is therefore staged: SHIFT-M3 should be used
as a \emph{first-pass} integrity filter whose outputs are routed to lightweight
verification rather than treated as automatic rejection decisions.
The method is equally applicable for post-market surveillance of deployed clinical decision support systems~\citep{kahn2012framework}: a screen that flags suspicious assembled records a posteriori can support ongoing monitoring of multimodal pipelines after deployment, not only at the point of record creation.
The next research step is not merely a larger encoder, but an identity-aware
consistency model that can distinguish ``different patient'' from ``same
patient, clinically changed.''
That future direction is well aligned with the informatics literature on data
quality and patient matching, where automated screening is most useful when it
feeds a broader verification workflow rather than acting as a fully autonomous
decision rule \citep{kahn2012framework,lewis2023ehr,gupta2024manual}.
The present study also has clear limitations: it relies on two text views,
uses TF-IDF rather than contextual encoders (though frozen pretrained encoders perform substantially worse in the present evaluation), evaluates on a single dataset (MEETI/MIMIC-IV-ECG) with only synthetic linkage errors, and shares a view-derivation dependency because the LLM interpretation is generated from the clinical report.
The auxiliary classification objective is disjoint from the identity
stream used for conflict detection. The reported results retain the evaluated
architecture, and conflict inference uses only the identity stream.
Conflict-score calibration is not evaluated in the present work; a calibrated score would allow direct mismatch-probability interpretation and is an explicit direction for future work.
The present evaluation is confined to a single dataset
(MEETI/MIMIC-IV-ECG), where the TF-IDF vocabulary, threshold calibration,
and stressor construction are all corpus-specific, and generalizability
to ECG recordings from other institutions or acquisition systems remains
to be established.

\paragraph*{Type-III as a representation-space stress test:}
The Type-III stressor is a controlled representation-space probe rather than a simulation of clinically realistic partial record corruption.
Real-world partial corruptions (including copied-forward documentation, split-record merges, and report amendments) manifest as coherent readable text that is internally consistent yet attributed to the wrong patient.
Type-III, by contrast, substitutes half of the TF-IDF feature vector with features drawn from a different patient, producing a hybrid representation with no readable-text analogue.
The reported \typeithree\% TPR should therefore be read as sensitivity to feature-level partial overlap under this controlled substitution, not as robustness to the full range of realistic partial record failures.

A more faithful approximation of readable partial corruption is the text-splice Type-III variant used in the masking analysis, in which the source interpretation's first half is concatenated with a donor interpretation's second half.
Under shared-token masking, this variant declines from AUROC \textspliceauroc\ to \textsplicemaskedauroc\ and from \textsplicetprfive\% to \textsplicetprfivemasked\% TPR@5\%, indicating that partial-swap detection is more dependent on lexical overlap than full-swap detection and that partial corruptions lacking shared surface tokens would require identity-aware signals beyond the scope of the present model.

\section{Conclusion}

This study frames linkage failure as a \emph{multimodal record integrity}
problem rather than a downstream prediction problem.
On MEETI, SHIFT-M3 shows that contrastively trained alignment between two separately
produced ECG text views is sufficient to detect full text-view swaps, partial swaps, and
hard negatives at strong operating points with a compact pre-fusion model.
Across same-dataset baselines, seed-stability analysis, loss ablation, and
shared-token masking, the evidence is consistent: the strongest signal comes
from cross-view contrastive alignment rather than lexical overlap alone.

The main limitation is longitudinal ambiguity.
Records from the same patient can change meaningfully over time, and the high
Type-II false-positive rate shows that content inconsistency is not yet the
same as identity mismatch.
For that reason, SHIFT-M3 is best viewed as a first-pass integrity screen
that prioritizes suspicious records for verification, not as a fully
autonomous adjudicator.
The next step is an identity-aware temporal consistency model that can separate
``different patient'' from ``same patient, clinically changed.''
More broadly, the results suggest that integrity checking should become a
standard stage before multimodal fusion in clinical AI pipelines, especially in
settings where record assembly errors can silently propagate downstream.


\bibliography{sample}

\newpage
\appendix
\section*{Appendix A: Replication Details}
\label{sec:appendix}

\subsection*{A.1\quad Dataset}
MEETI is publicly available via PhysioNet~\citep{meeti2026}.
Records are split 80/10/10 at the subject level, assigning all
visits of a given patient to the same partition
(Table~\ref{tab:dataset_miccai}).

\subsection*{A.2\quad Feature Extraction}
Separate unigram TF-IDF vocabularies are fit on the training split
only, using 500 interpretation features and 325 report features.
Standard English stop words are removed, and the vocabularies are
applied to validation and test sets without refitting.
Feature vectors are L2-normalized before input to the MLP streams.

\subsection*{A.3\quad Architecture}
Each identity MLP stream follows Linear($d_{in}$, 256) $\to$ ReLU
$\to$ BatchNorm $\to$ Dropout(0.2) $\to$ Linear(256, 128), with
$d_{in}{=}500$ for the interpretation stream and $d_{in}{=}325$
for the report stream. Its output embeddings are L2-normalized. The auxiliary
task stream uses parallel modality-specific projections
($d_{in}\!\to\!256\!\to\!128$), followed by a classifier
($256\!\to\!64\!\to\!1$).
The task stream shares no parameters with the identity stream and is not used
at inference: the conflict score is computed solely from the identity-stream
embeddings, and the loss ablation in Table~\ref{tab:e29_miccai} confirms that
zeroing the task loss leaves every conflict metric unchanged.
It is retained here only to describe the architecture as trained.
Total parameters: 573,569 including the auxiliary task stream, of which only
the identity stream is active at screening time.

\subsection*{A.4\quad Training Configuration}
Table~\ref{tab:hparams} lists all hyperparameters.
Three random seeds (42, 43, 44) are used for the stability analysis
reported in Table~\ref{tab:e28_perseed_miccai}.
All experiments were run on a single NVIDIA Tesla T4 GPU (16\,GB).

\begin{table}[h]
\centering
\caption{Hyperparameter configuration used for all reported experiments.}
\label{tab:hparams}
\small
\begin{tabular}{ll}
\toprule
\textbf{Hyperparameter} & \textbf{Value} \\
\midrule
Optimizer          & Adam \\
$\beta_1$ / $\beta_2$ / $\epsilon$ & 0.9 / 0.999 / $10^{-8}$ \\
Learning rate      & $3 \times 10^{-4}$ \\
Batch size         & 128 \\
Epochs             & 10 \\
Embedding dim      & 128 \\
$\lambda_{ctr}$    & 1.0 \\
$\lambda_{temp}$   & 0.5 \\
Swap margin $m$    & 0.5 \\
Dropout            & 0.2 \\
Random seeds       & 42, 43, 44 \\
\bottomrule
\end{tabular}
\end{table}

\subsection*{A.5\quad Operating-Point Threshold}
The default 5\% FPR operating point corresponds to the 95th
percentile of clean-pair conflict scores on the validation split.
The frontier in Table~\ref{tab:frontier_miccai} sweeps from the
99th percentile (1\% FPR) to the 80th (20\% FPR).

\subsection*{A.6\quad Stressor Construction}
All stressors are constructed from held-out test records; no
stressor pair appears in training.
\begin{itemize}[noitemsep,topsep=2pt]
  \item \emph{Type-I:} $t_B$ is replaced by a randomly drawn
        different-subject report.
  \item \emph{Type-II:} two records from the same subject at
        different visits are paired.
  \item \emph{Type-III:} 250 of 500 interpretation feature
        dimensions are replaced by the corresponding dimensions
        from a randomly drawn different-subject interpretation
        (feature-space substitution, not text splicing).
  \item \emph{Hard Negative:} each record is paired with a
        different-subject record sharing the same coarse keyword
        label (normal vs.\ abnormal).
  \item \emph{Text-splice Type-III} (masking analysis only):
        the first half of the source interpretation text is
        concatenated with the second half of a donor
        interpretation.
\end{itemize}

\subsection*{A.7\quad Shared-Token Masking}
The shared-token mask is computed as the intersection of non-zero
vocabulary indices in $t_A$ and $t_B$. Those indices are zeroed
in both feature vectors before conflict-score computation.

\end{document}